\documentclass{article}

\usepackage[utf8]{inputenc}
\usepackage[T1]{fontenc}
\usepackage[english]{babel}

\usepackage{graphicx}
\usepackage{booktabs}
\usepackage{geometry}
\usepackage{amsmath}
\usepackage{float}
\usepackage{url}
\usepackage{hyperref}
\usepackage{subcaption} 
\usepackage{xcolor}

\newcommand{\datasetversion}{v1.0.0} 
\newcommand{\datasetdoi}{10.5281/zenodo.18189192} 
\newcommand{\dataseturl}{https://doi.org/\datasetdoi}
\newcommand{\codeurl}{https://github.com/lucMorello/Dengue-Brasil-Arboviroses-Dataset-Brazil-Dengue-Arboviral-Diseases-Dataset.git} 
\newcommand{\licensepaper}{arXiv perpetual, non-exclusive license} 
\newcommand{\licensedata}{CC BY 4.0} 

\title{ A Dataset of Dengue Hospitalizations in  Brazil (1999–2021) with Weekly Disaggregation from Monthly Counts}
\author{
  Lucas M. Morello$^{1}$$^{,3}$, Matheus Lima Castro$^{1}$$^{,3}$$^{,4}$, Pedro Cesar M. G. Camargo$^{1}$$^{,4}$, \\ Liliane Moreira Nery$^{2}$, Darllan Collins da Cunha e Silva$^{2}$, Leopoldo Lusquino Filho$^{1}$$^{,4}$ \\
  {\small $^{1}$Dept. of Control and Automation Engineering, UNESP}\\
  {\small $^{2}$Dept. of  Environmental Science, UNESP}
  \\ 
  {\small $^{3}$ HairU Servi\c{c}os e Tecnologia LTDA } \\
  {\small $^{4}$ recod.ai, UNICAMP } \\
  {\small \texttt{lucasmmorello@gmail.com}}
}
\date{} 

\begin{document}
\maketitle

\begin{abstract}
This data paper describes and publicly releases these dataset  (\datasetversion), published on Zenodo under DOI \href{\dataseturl}{\datasetdoi}. Motivated by the need to increase the temporal granularity of originally monthly data to enable more effective training of AI models for epidemiological forecasting, the dataset harmonizes municipal-level dengue hospitalization time series across Brazil and disaggregates them to weekly resolution (epidemiological weeks) through an interpolation protocol with a correction step that preserves monthly totals. The statistical and temporal validity of this disaggregation was assessed using a high-resolution reference dataset from the state of S\~ao Paulo (2024), which simultaneously provides monthly and epidemiological-week counts, enabling a direct comparison of three strategies: linear interpolation, \textit{jittering}, and \textit{cubic spline}. Results indicated that \textit{cubic spline} interpolation achieved the highest adherence to the reference data, and this strategy was therefore adopted to generate weekly series for the 1999--2021 period. In addition to hospitalization time series, the dataset includes a comprehensive set of explanatory variables commonly used in epidemiological and environmental modeling, such as demographic density, CH$_4$, CO$_2$, and NO$_2$ emissions, poverty and urbanization indices, maximum temperature, mean monthly precipitation, minimum relative humidity, and municipal latitude and longitude, following the same temporal disaggregation scheme to ensure multivariate compatibility. The paper documents the dataset’s provenance, structure, formats, licenses, limitations, and quality metrics (MAE, RMSE, $R^2$, KL, JSD, DTW, and the KS test), and provides usage recommendations for multivariate time-series analysis, environmental health studies, and the development of \textit{machine learning} and \textit{deep learning} models for outbreak forecasting.
\end{abstract}

\noindent\textbf{Keywords:} dengue; hospitalizations; epidemiological surveillance; time series; temporal disaggregation; interpolation; dataset; Zenodo; DOI.

\section*{Data availability and License}
\textbf{Dataset version:}  (\datasetversion) available at\href{\dataseturl}{\datasetdoi}.  
\textbf{Dataset license:} \licensedata.   
\textbf{Code:} \href{\codeurl}{\codeurl}.  

\section{Introduction}

Dengue hospitalization data were obtained from DataSUS and are originally reported at a monthly resolution. The temporal coverage spans 1999 to 2021, yielding time series with 264 observations per municipality. Because the literature highlights the difficulty of training \textit{Machine Learning} models on short time series \cite{Interpolation_Time_Series}, this work adopts temporal disaggregation via interpolation. The goal is to expand the dataset to support predictive modeling and analysis at weekly granularity (epidemiological weeks), while ensuring reproducibility and DOI-based versioning \cite{better_perform}.

To increase temporal resolution, two main approaches were considered: Generative Adversarial Networks (GANs) and mathematical interpolation. The former relies on training a generative model to learn data trends and produce synthetic samples. However, GAN-based augmentation was deemed impractical in this context, since municipal time series are highly heterogeneous, exhibiting distinct peaks and municipality-specific distortions that hinder the training of a single unified model. Moreover, a length of 264 points is typically insufficient for stable GAN training.

Therefore, interpolation was selected to generate additional time points. This choice is also motivated by its practical relevance to public health management, where epidemiological weeks are the standard temporal unit. Three interpolation strategies were compared: linear interpolation, \textit{cubic spline}, and \textit{jittering}. Importantly, the procedure enforces two constraints: (i) the sum of weekly values within each month matches the corresponding observed monthly total, and (ii) each month is assigned the correct number of epidemiological weeks associated with that month.


\section{Methodology}
Dengue hospitalization records were obtained from DataSUS \cite{datasus2025}, an official Brazilian government repository linked to the Unified Health System (SUS). Complementary explanatory variables were collected from the Brazilian Institute of Geography and Statistics (IBGE) \cite{ibge2025}. An initial cleaning step removed municipalities with missing values in the variables used. To mitigate scale effects across municipalities with different population sizes, the outcome of interest was expressed as a hospitalization rate, defined as the ratio between the number of dengue hospitalizations and the municipal population.

To identify the interpolation strategy that best reproduces the real-world dynamics of dengue hospitalizations, we relied on a reference dataset released by the Center for Epidemiological Surveillance ``Prof.\ Alexandre Vranjac'' (CVE-SP) \cite{Dengue_cases_in_SP}. This dataset is central to the validation protocol because it provides São Paulo state hospitalization counts for 2024 simultaneously at monthly resolution and at epidemiological-week resolution (using autochthonous dengue cases, i.e., cases among residents of the same municipality). Epidemiological weeks are the standard temporal unit in public health surveillance and follow an international convention that partitions the calendar into consecutive seven-day intervals starting on Sundays, facilitating monitoring and comparison of outbreaks across civil months.

The validation procedure, illustrated in Figure \ref{fig:pipeline}, consisted of applying the three candidate interpolation methods, Linear, \textit{Cubic Spline}, and \textit{Jittering}, to the observed monthly series and comparing the resulting synthetic weekly series against the weekly series observed in CVE-SP. Although epidemiological weeks do not align perfectly with civil months, we assume that the resulting temporal offset is negligible for the purposes of method comparison. Temporal disaggregation was implemented in Python using \textit{Pandas}, \textit{NumPy}, and SciPy \cite{SciPyKS2samp}, enforcing two key constraints: the correct number of epidemiological weeks assigned to each month and exact preservation of the monthly totals after disaggregation.

After generating the synthetic weekly series, each method was evaluated by direct comparison to the CVE-SP weekly reference. For analysis, performance measures were organized into two complementary groups. The first group comprises pointwise error metrics (MAE, RMSE, and $R^2$), which quantify numerical agreement between interpolated and observed weekly values. The second group comprises distributional and temporal similarity metrics, including the Kullback--Leibler and Jensen--Shannon divergences, as well as \textit{Dynamic Time Warping} (DTW), to assess whether the interpolation preserves both the probability distribution and the temporal structure of outbreak dynamics.

\section{Dataset Construction: Monthly-to-Weekly Temporal Disaggregation}

The temporal disaggregation to weekly resolution (epidemiological weeks) was carried out under two primary constraints for the dengue hospitalization target variable: (i) the number of weeks assigned to each civil month must match the actual number of epidemiological weeks that intersect that month; and (ii) the sum of the weekly values assigned to a given month must be exactly equal to the original monthly value, ensuring conservation of hospitalizations and preventing distortions in the monthly totals. In other words, the dengue disaggregation strictly preserves the monthly aggregates while redistributing them across the corresponding epidemiological weeks.

For the explanatory variables (e.g., demographic density index, CH$_4$, CO$_2$, and NO$_2$ emissions, poverty index, urbanization index, maximum temperature, mean monthly precipitation, and minimum relative humidity), the monthly-sum conservation constraint is not appropriate because these covariates are not additive counts that should be partitioned across weeks. Therefore, for these covariates we enforced only constraint (i): each monthly value is propagated to the epidemiological weeks belonging to the respective month, preserving temporal coherence and ensuring consistent alignment between the derived weekly covariates and the dengue target series. This design maintains the calendar synchronization between dengue and covariates at weekly resolution without imposing an artificial conservation-of-sum requirement on non-additive quantities.

Accordingly, three interpolation strategies were selected to assess which yields the highest statistical adherence in reconstructing weekly dengue hospitalizations.

\subsection{Linear Interpolation}
This approach uniformly distributes each month’s total number of dengue hospitalizations across the epidemiological weeks associated with that month. To preserve fidelity to the original monthly aggregates, a numerical correction factor is applied so that the sum of the resulting weekly values matches the observed monthly total exactly, yielding a step-like weekly profile.

Monthly-sum preservation is enforced through a correction factor $\phi$. For a month with total value $V_m$ and $n_m$ corresponding epidemiological weeks, the corrected weekly value $w_{j}^{(\mathrm{final})}$ is obtained as:
\begin{equation}
\label{eq:linear_interpolation_factor}
\phi = \frac{V_m}{\sum_{j=1}^{n_m} w_{j}^{(\mathrm{base})}}
\quad , \quad
w_{j}^{(\mathrm{final})} = w_{j}^{(\mathrm{base})}\cdot \phi.
\end{equation}

This method is used as the \textit{baseline} because it assumes intra-month constancy and preserves only the monthly totals.

\subsection{Jittering Interpolation}
Jittering-based interpolation adds random noise to the reconstructed weekly series in order to introduce local variability while preserving the overall structure of the data \cite{interpolation_jittering}. In this work, the procedure follows the same initial step as linear interpolation: each monthly total is first uniformly redistributed across the epidemiological weeks associated with that month, producing a step-like baseline profile.

Unlike pure linear interpolation, however, an additive Gaussian noise term with zero mean is injected at the weekly level. Specifically, for each week $j$ within month $m$, we add a perturbation $\epsilon_j$ drawn from a normal distribution whose standard deviation is proportional to $0.05$ times the mean weekly value of that month.

To ensure epidemiological plausibility, two constraints are enforced: (i) non-negativity of weekly values, implemented by rejecting or rectifying negative realizations (clipping to zero), and (ii) exact preservation of the monthly total, enforced via a multiplicative correction applied after noise injection. The resulting formulation is:
\begin{equation}
\label{eq:jittering_interpolation_factor}
w_{j}^{(\mathrm{base})} = \left(\frac{V_m}{n_m}\right) + \epsilon_j
\quad , \quad
w_{j}^{(\mathrm{final})} = w_{j}^{(\mathrm{base})}\cdot
\left(\frac{V_m}{\sum_{j=1}^{n_m} w_{j}^{(\mathrm{base})}}\right).
\end{equation}

\subsection{Cubic Spline Interpolation}
Cubic spline interpolation aims to estimate intermediate values between discrete observations using third-order polynomial functions, ensuring smoothness and continuity across the reconstructed curve. In this formulation, each pair of consecutive points is connected by a cubic polynomial
$s_i(x)=a_i x^3 + b_i x^2 + c_i x + d_i$,
with constraints that enforce continuity of the function and its first and second derivatives at the knots. To obtain the interpolated trajectory, we used the \textit{CubicSpline} implementation from the \textit{SciPy} library \cite{CubicSpline}.

Because cubic splines require boundary conditions and the series does not provide additional information at the extremities, we adopted a practical stabilization strategy: the first monthly value was duplicated prior to spline fitting to reduce edge effects. After interpolation, this auxiliary point was removed so that the reconstructed series preserves the original temporal support.

In the monthly-to-weekly conversion, the cubic spline was fitted over the monthly observations to generate a smooth underlying curve, from which intermediate values were sampled to match the epidemiological-week grid. The resulting weekly sequence yields preliminary values $w_{j}^{(\text{base})}$, which are then allocated according to the true number of epidemiological weeks associated with each calendar month. Finally, a multiplicative correction factor is applied so that, for the dengue hospitalization target, the sum of the weekly values within each month exactly matches the original monthly total, ensuring strict monthly-mass conservation.

\section{Quality Control and Validation}
\subsection{Reference Dataset (CVE-SP 2024)}
To validate the fidelity of the monthly-to-weekly disaggregation, we used the CVE-SP dataset \cite{Dengue_cases_in_SP}. This reference set contains 2024 records for the state of São Paulo with both monthly aggregates and epidemiological-week counts (autochthonous cases), enabling a direct, paired comparison.

\begin{figure}[H]
\centering
\includegraphics[width=0.85\textwidth]{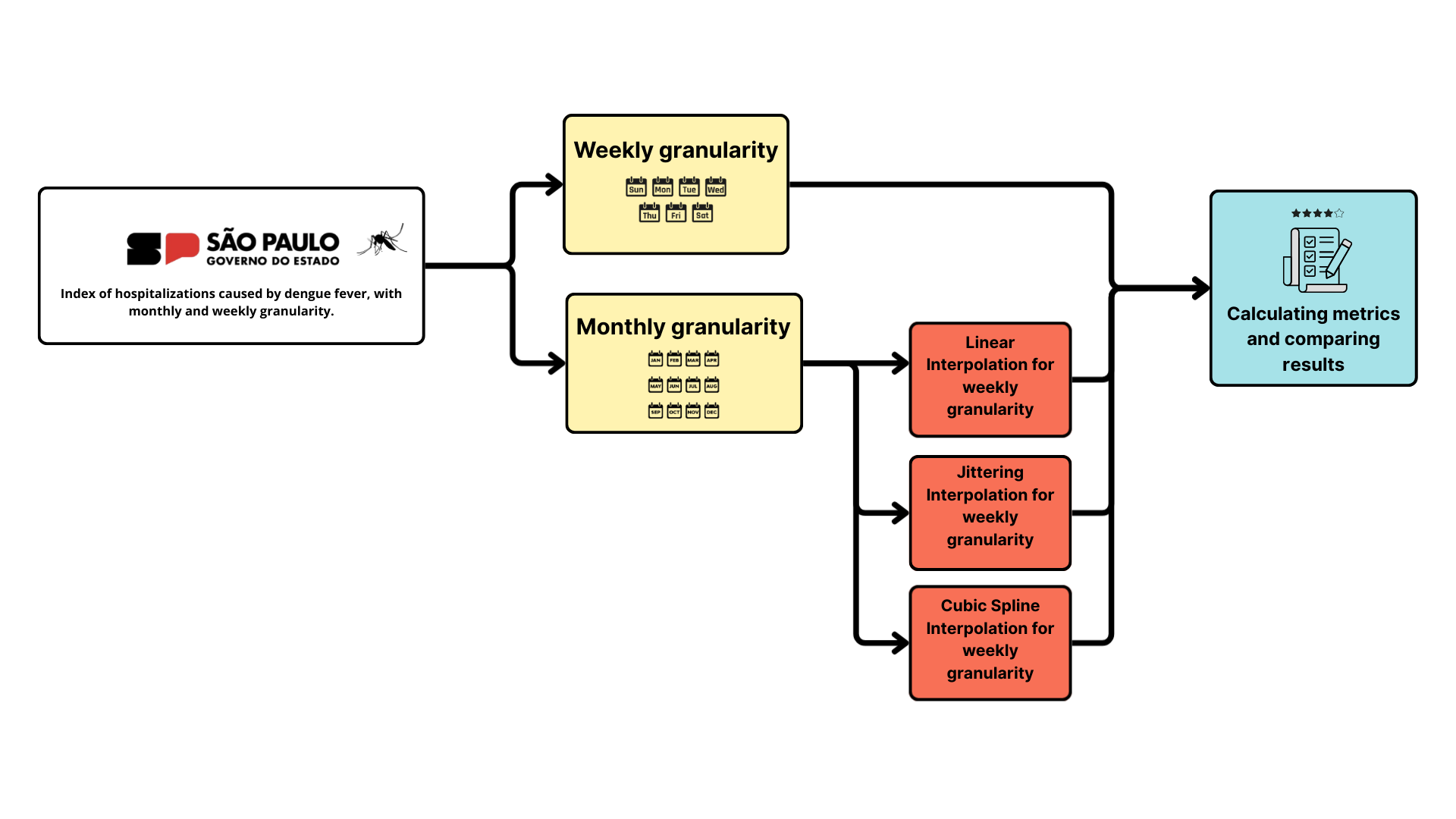}
\caption{Validation pipeline used to compare the interpolation methodologies.}
\label{fig:pipeline}
\end{figure}

The validation procedure, illustrated in Figure \ref{fig:pipeline}, consisted of applying the three proposed interpolation strategies (Linear, \textit{Cubic Spline}, and \textit{Jittering}) to the real monthly series and then comparing the resulting synthetic weekly series against the observed weekly series from CVE-SP. Although epidemiological weeks do not align perfectly with civil calendar months, we assume that any boundary misalignment is small enough for the purposes of methodological comparison. The entire pipeline was implemented in Python using \textit{Pandas}, \textit{NumPy}, and SciPy \cite{SciPyKS2samp}.

\subsection{Evaluated Metrics}
Due to the idiosyncrasies of each time series, there is no widely accepted consensus in the literature regarding an optimal set of metrics for assessing interpolation quality and the validity of temporally disaggregated data \cite{metrics_data_argumentation}. Therefore, we adopted two complementary groups of criteria: pointwise error metrics (MAE, RMSE, and $R^2$) and distributional/temporal similarity metrics (KL, JSD, DTW, and the Kolmogorov--Smirnov test).

\subsubsection{Mean Absolute Error (MAE)}
\begin{equation}
    \label{eq:MAE}
    MAE = \frac{1}{n} \sum_{i = 1}^{n} \left|X_i - Y_i\right|
\end{equation}

\subsubsection{Root Mean Squared Error (RMSE)}
\begin{equation}
    \label{eq:RMSE}
    RMSE = \sqrt{\frac{1}{n}\sum_{i = 1}^{n}\left(X_i - Y_i\right)^2}
\end{equation}

\subsubsection{Coefficient of Determination}
\begin{equation}
    \label{eq:R2}
    R^2 = 1 - \frac{\sum_{i=1}^{n} (Y_i - X_i)^2}{\sum_{i=1}^{n} (Y_i - \bar{Y})^2}
\end{equation}

\subsubsection{Kullback--Leibler Divergence}
\begin{equation}
    \label{eq:D_kl}
    D_{\text{KL}}(p \parallel q) = \sum_{x \in \mathcal{X}} p(x)\,\ln\left(\frac{p(x)}{q(x)}\right)
\end{equation}

\subsubsection{Jensen--Shannon Divergence}
\begin{equation}
    \label{eq:media_distribuicoes}
    m(x) = \frac{1}{2}\left(p(x) + q(x)\right)
\end{equation}

\begin{equation}
    \label{eq:JSD}
    \mathrm{JSD}(p \parallel q) =
    \frac{1}{2} D_{\mathrm{KL}}(p \parallel m)
    + \frac{1}{2} D_{\mathrm{KL}}(q \parallel m)
\end{equation}

\subsubsection{Dynamic Time Warping (DTW)}
Dynamic Time Warping (DTW) compares time series by allowing elastic temporal alignments, which is useful when similar epidemic patterns may be phase-shifted in time \cite{DTW}. Because DTW is unbounded above and depends on series length, we also report a normalized DTW variant.

\subsubsection{Kolmogorov--Smirnov Test}
The Kolmogorov--Smirnov (KS) test assesses whether two samples are drawn from the same distribution by computing the maximum distance between their empirical cumulative distribution functions (ECDFs) \cite{KS_teste,SciPy_kstest}.

\subsection{KDE and Bandwidth Selection (Silverman's Rule)}
The computation of KL and JSD relies on kernel density estimation (KDE). To ensure comparability across interpolation methods, we fixed the KDE bandwidth on a per-municipality basis using Silverman’s rule of thumb computed from the \emph{observed} (reference) weekly series \cite{Silverman_livro}. When the empirical variability was null (e.g., constant or near-constant series), a minimum bandwidth of $0.1$ was enforced to avoid numerical degeneracy:
\begin{equation}
\label{eq:silverman}
\hat{b} = 1.06 \cdot \min\left( \hat{\sigma}, \frac{\mathrm{IQR}}{1.34} \right) \cdot n^{-1/5}.
\end{equation}

For the Kolmogorov--Smirnov test, we consider the following hypotheses:
\begin{itemize}
    \item \textbf{Null hypothesis ($H_0$):} $p(x)=q(x)$.
    \item \textbf{Alternative hypothesis ($H_1$):} $p(x)\neq q(x)$.
\end{itemize}

Finally, we note that the sMAPE metric was excluded due to numerical instability in highly sparse series with a large proportion of zeros.

\section{Results}
The three interpolation techniques were applied to all 645 municipalities in the state of São Paulo. Given the scale of the dataset, results are primarily reported in an aggregated form through summary tables of global metrics and \textit{boxplots}, enabling an assessment of the statistical distribution of each method’s performance across municipalities. In addition, we provide municipality-level visualizations for selected cities to support a qualitative inspection of the interpolated time-series behavior under distinct epidemiological profiles.

\subsection{Municipality-level Plots}
\begin{figure}[H]
\centering
\includegraphics[width=0.85\textwidth]{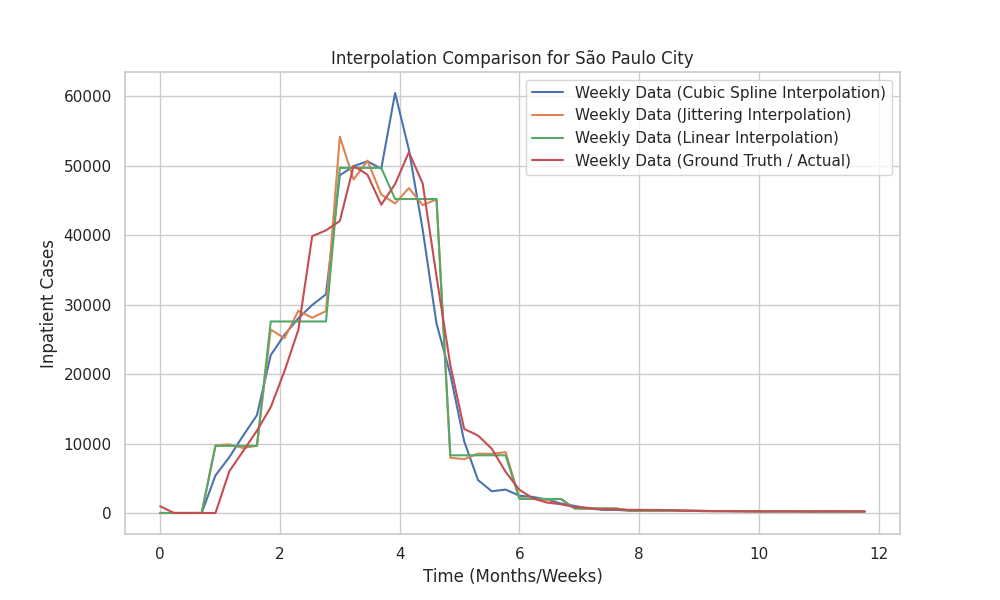}
\caption{Interpolation results for the municipality of São Paulo.}
\label{fig:city_SP_dengue}
\end{figure}

For a qualitative assessment of the time series, we selected four municipalities representing distinct and challenging scenarios. São Paulo was chosen because it exhibits the largest absolute number of cases (Figure \ref{fig:city_SP_dengue}), serving as a stability benchmark for interpolation methods under high-incidence, high-volume dynamics. Braúna was selected because it yielded the lowest mean $R^2$ across all evaluated methods, enabling the identification of methodological limitations under low-adherence and highly sparse regimes. In contrast, Suzano was included for achieving the highest global mean $R^2$, whereas Arujá was highlighted for presenting the best fit under the \textit{Cubic Spline} approach.

The visual analysis of São Paulo (Figure \ref{fig:city_SP_dengue}) indicates a high overall agreement between the interpolated series and the observed weekly reference. Although the \textit{Cubic Spline} method exhibits stronger local instability—manifested as more pronounced oscillations (peaks and troughs), particularly around the transitions between months 4, 5, and 6—it provides the most faithful representation of the global dynamics and curvature of the observed trajectory. This suggests that, despite pointwise fluctuations induced by the polynomial smoothing, the cubic model better captures the non-linear acceleration and deceleration phases typical of epidemic surges in high-incidence settings.

\begin{figure}[H]
\centering
\includegraphics[width=0.85\textwidth]{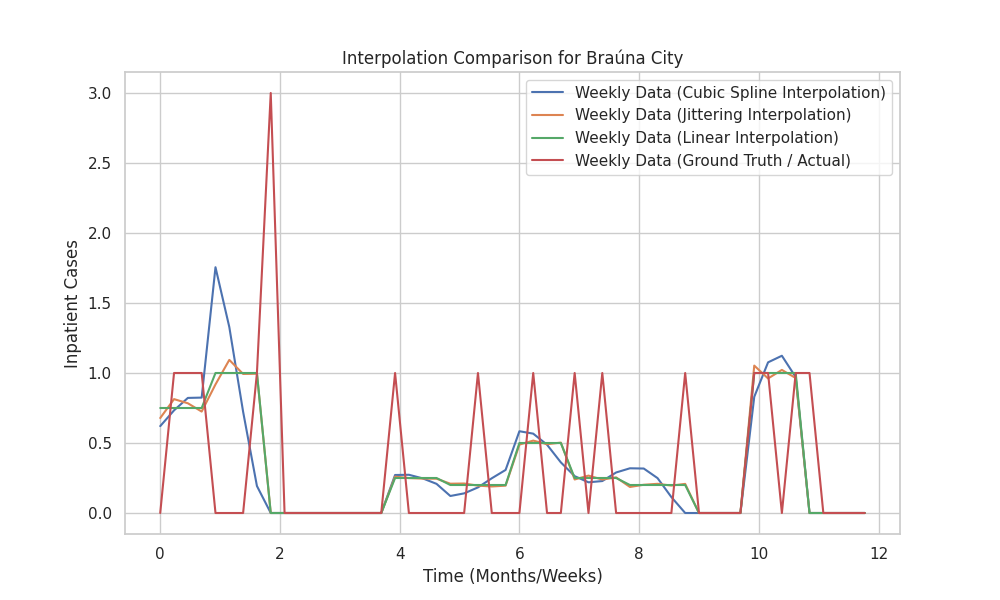}
\caption{Interpolation results for the municipality of Braúna.}
\label{fig:city_Branauna_dengue}
\end{figure}

Braúna (Figure \ref{fig:city_Branauna_dengue}) exhibited the lowest performance in terms of $R^2$ among the analyzed municipalities. This outcome is primarily explained by the extreme sparsity and high volatility of the original time series. The weekly reference displays an intermittent pattern, in which isolated weeks with small counts are followed by extended periods with zero hospitalizations. Because the interpolation procedures operate on monthly aggregates, the lower temporal resolution inherently smooths or redistributes these rare events and cannot reproduce the discrete, punctual fluctuations observed in the weekly ground truth, leading to an unavoidable loss of information and reduced goodness-of-fit in low-incidence regimes.

\begin{figure}[H]
\centering
\includegraphics[width=0.85\textwidth]{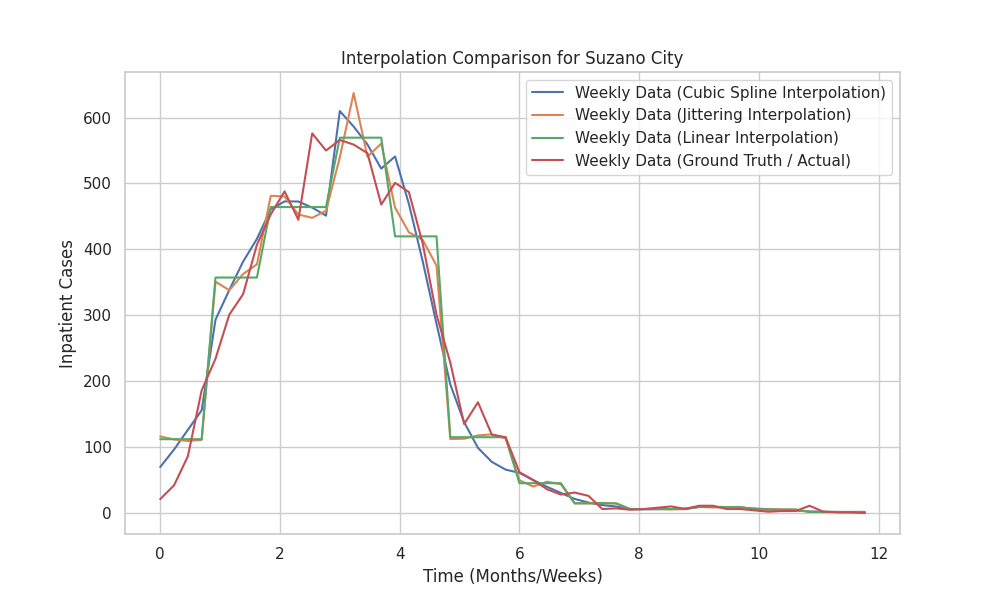}
\caption{Interpolation results for the municipality of Suzano.}
\label{fig:city_Suzano_dengue}
\end{figure}

Suzano (Figure \ref{fig:city_Suzano_dengue}) achieved the highest average $R^2$. Its temporal profile is approximately unimodal (bell-shaped), with a well-defined outbreak and comparatively smooth transitions over the months, without abrupt discontinuities or excessive noise. This regularity likely favors interpolation performance, because downscaling from monthly to weekly resolution does not discard substantial high-frequency structure; instead, the interpolated series can recover the central trend and outbreak morphology with high fidelity.

\begin{figure}[H]
\centering
\includegraphics[width=0.85\textwidth]{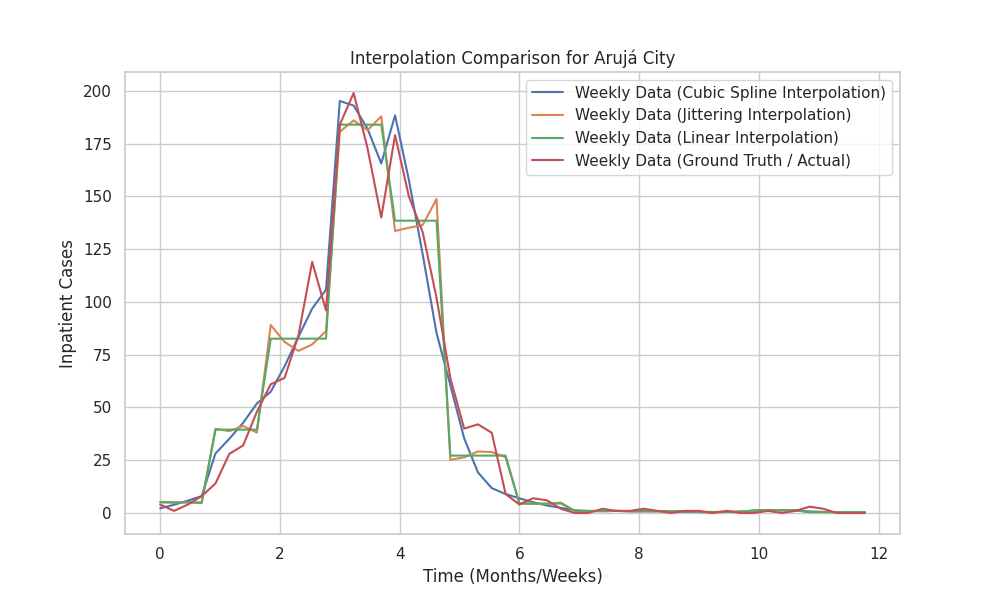}
\caption{Interpolation results for the municipality of Arujá.}
\label{fig:city_Arujá_dengue}
\end{figure}

Arujá (Figure \ref{fig:city_Arujá_dengue}) yielded the highest $R^2$ under the \textit{Cubic Spline} method. Visual inspection suggests that this superior fit is associated with the presence of stronger dynamics and more pronounced epidemic peaks. Unlike the Linear and \textit{Jittering} approaches, which produce step-like intra-month profiles, cubic splines rely on third-order polynomials that provide greater sensitivity to changes in curvature, enabling a smoother and more realistic reconstruction of acceleration and deceleration phases of the outbreak trajectory.

\subsection{Error-based metrics}
The results for the error-based metrics were organized into two tables. Table \ref{tab:metrics_interpolation_mean} reports aggregated descriptive statistics across all 645 municipalities in the state of S\~ao Paulo. In this setting, absolute error metrics (MAE and RMSE) are directly influenced by population size and the absolute volume of hospitalizations, leading to a scale bias in which large municipalities disproportionately increase the global mean. To mitigate this effect and enable a more qualitative comparison across distinct epidemiological profiles, Table \ref{tab:metrics_selected_municipalities} presents municipality-level results for the representative cases discussed earlier (S\~ao Paulo, Bra\'una, Suzano, and Aruj\'a).

\begin{table}[htbp]
\centering
\caption{Descriptive Statistics of Error Metrics for the Interpolation Methods (n=645)}
\label{tab:metrics_interpolation_mean}
\resizebox{\textwidth}{!}{%
\begin{tabular}{@{}lccccccccc@{}}
\toprule
\textbf{Statistic} & \multicolumn{3}{c}{\textbf{Linear}} & \multicolumn{3}{c}{\textbf{Cubic Spline}} & \multicolumn{3}{c}{\textbf{Jittering}} \\ \cmidrule(lr){2-4} \cmidrule(lr){5-7} \cmidrule(lr){8-10}
 & \textit{MAE} & \textit{RMSE} & \textit{$R^2$} & \textit{MAE} & \textit{RMSE} & \textit{$R^2$} & \textit{MAE} & \textit{RMSE} & \textit{$R^2$} \\ \midrule
Mean & 13.94 & 25.53 & \textbf{0.711} & \textbf{12.29} & \textbf{22.79} & 0.698 & 14.46 & 26.36 & 0.708 \\
Std. Dev. & 100.21 & 189.01 & 0.229 & \textbf{85.13} & \textbf{154.88} & 0.255 & 108.76 & 202.03 & \textbf{0.228} \\
Minimum & \textbf{0.03} & \textbf{0.12} & \textbf{-0.109} & 0.03 & 0.13 & -0.310 & 0.03 & 0.12 & -0.102 \\
25\% & \textbf{0.94} & \textbf{1.62} & \textbf{0.595} & 0.97 & 1.69 & 0.552 & 0.95 & 1.63 & 0.587 \\
Median (50\%) & 2.42 & \textbf{3.89} & \textbf{0.792} & \textbf{2.24} & 4.08 & \textbf{0.792} & 2.43 & 3.97 & 0.785 \\
75\% & 7.64 & 13.61 & 0.891 & \textbf{7.00} & \textbf{12.79} & \textbf{0.898} & 7.65 & 13.74 & 0.889 \\
Maximum & 2432.20 & 4613.02 & 0.967 & \textbf{2071.32} & \textbf{3771.54} & \textbf{0.979} & 2654.83 & 4950.32 & 0.966 \\ \bottomrule
\end{tabular}
}%
\subcaption*{\small Note: Bold values indicate the best performance for each metric.}
\end{table}

\begin{table}[htbp]
\centering
\caption{Comparison of Error Metrics for Selected Municipalities}
\label{tab:metrics_selected_municipalities}
\resizebox{\textwidth}{!}{%
\begin{tabular}{@{}lccccccccc@{}}
\toprule
\textbf{Municipality} & \multicolumn{3}{c}{\textbf{Linear}} & \multicolumn{3}{c}{\textbf{Cubic Spline}} & \multicolumn{3}{c}{\textbf{Jittering}} \\ \cmidrule(lr){2-4} \cmidrule(lr){5-7} \cmidrule(lr){8-10}
 & \textit{MAE} & \textit{RMSE} & \textit{$R^2$} & \textit{MAE} & \textit{RMSE} & \textit{$R^2$} & \textit{MAE} & \textit{RMSE} & \textit{$R^2$} \\ \midrule
Aruj\'a    & 8.34 & 14.47 & 0.938 & \textbf{5.02} & \textbf{8.48} & \textbf{0.979} & 8.44 & 14.56 & 0.938 \\
Bra\'una   & \textbf{0.35} & \textbf{0.61} & \textbf{-0.109} & 0.39 & 0.66 & -0.310 & 0.35 & 0.61 & -0.102 \\
S\~ao Paulo & 2432.20 & 4613.02 & 0.928 & \textbf{2071.32} & \textbf{3771.54} & \textbf{0.952} & 2654.83 & 4950.32 & 0.918 \\
Suzano     & 28.94 & 47.11 & 0.947 & \textbf{21.09} & \textbf{33.36} & \textbf{0.974} & 30.53 & 48.10 & 0.945 \\ \bottomrule
\end{tabular}%
}
\subcaption*{\small Note: Bold values indicate the best performance for each municipality.}
\end{table}

When comparing the \textit{Linear} and \textit{Cubic Spline} methods, average performance across S\~ao Paulo is broadly similar, but clear differences emerge depending on the epidemiological profile. \textit{Cubic Spline} tends to perform better in municipalities with higher case density, as it more accurately captures outbreak acceleration, whereas in sparse series with many zeros it may introduce instabilities, making the linear approach more robust. The \textit{Jittering} strategy did not yield statistically meaningful improvements, suggesting that weekly variability is not well represented by Gaussian noise and that discrepancies are more likely driven by structural reporting patterns or notification delays. The error metrics also exhibit strong asymmetry (mean $>$ median), indicating a disproportionate influence of large urban centers on aggregated statistics; moreover, the minimum $R^2$ values highlight the limitations of higher-order polynomial methods under extreme sparsity. Therefore, selecting a single ``best'' method should not rely solely on pointwise error metrics, motivating complementary evaluation using statistical similarity measures.

\subsection{Statistical similarity metrics}

The results for the distributional similarity and temporal distance metrics (KL, JSD, DTW, and the KS test \textit{p-value}) were organized into two tables. Table \ref{tab:advanced_statistical_metrics} summarizes descriptive statistics across the 645 municipalities of S\~ao Paulo, whereas Table \ref{tab:advanced_municipal_metrics} reports the results for the selected case-study cities (S\~ao Paulo, Bra\'una, Suzano, and Aruj\'a), enabling a comparison of the interpolation strategies under distinct epidemiological profiles.

\begin{table}[htbp]
\centering
\caption{Statistical Metrics of Distributional Similarity and Temporal Distance (n=645)}
\label{tab:advanced_statistical_metrics}
\resizebox{\textwidth}{!}{%
\begin{tabular}{@{}lcccccccccccc@{}}
\toprule
\textbf{Statistic} & \multicolumn{3}{c}{\textbf{KL Divergence} ($\downarrow$)} & \multicolumn{3}{c}{\textbf{JSD Distance} ($\downarrow$)} & \multicolumn{3}{c}{\textbf{DTW Distance} ($\downarrow$)} & \multicolumn{3}{c}{\textbf{p-value} ($\uparrow$)} \\ \cmidrule(lr){2-4} \cmidrule(lr){5-7} \cmidrule(lr){8-10} \cmidrule(lr){11-13}
 & \textit{Lin.} & \textit{Cub.} & \textit{Jit.} & \textit{Lin.} & \textit{Cub.} & \textit{Jit.} & \textit{Lin.} & \textit{Cub.} & \textit{Jit.} & \textit{Lin.} & \textit{Cub.} & \textit{Jit.} \\ \midrule
Mean & inf & \textbf{0.141} & inf & 0.163 & \textbf{0.121} & 0.156 & 616.51 & \textbf{445.86} & 589.76 & 0.424 & \textbf{0.508} & 0.428 \\
Std. Dev. & -- & 1.161 & -- & 0.088 & \textbf{0.070} & 0.087 & 4252.52 & \textbf{3064.87} & 3929.54 & 0.348 & 0.380 & 0.357 \\
Minimum & 0.00 & 0.00 & 0.00 & 0.00 & 0.00 & 0.00 & 1.50 & \textbf{1.19} & 1.47 & 0.000 & 0.000 & 0.000 \\
25\% & 0.036 & \textbf{0.018} & 0.029 & 0.096 & \textbf{0.067} & 0.084 & 41.60 & \textbf{38.40} & 40.98 & 0.046 & \textbf{0.126} & 0.046 \\
Median & 0.106 & \textbf{0.049} & 0.091 & 0.157 & \textbf{0.111} & 0.148 & 101.65 & \textbf{87.28} & 101.21 & 0.421 & \textbf{0.574} & 0.421 \\
75\% & 0.239 & \textbf{0.098} & 0.213 & 0.220 & \textbf{0.160} & 0.213 & 317.65 & \textbf{244.79} & 308.15 & 0.739 & \textbf{0.884} & 0.739 \\
Maximum & inf & 27.99 & inf & 0.438 & \textbf{0.385} & 0.437 & 102481.4 & \textbf{74075.8} & 94147.4 & 1.000 & \textbf{1.000} & 1.000 \\ \bottomrule
\end{tabular}%
}
\subcaption*{\small Note: ($\downarrow$) indicates that lower values are better; ($\uparrow$) indicates that higher values are better. KL: Kullback--Leibler; JSD: Jensen--Shannon; DTW: Dynamic Time Warping.}
\end{table}

\begin{figure}[H]
\centering
\includegraphics[width=0.85\textwidth]{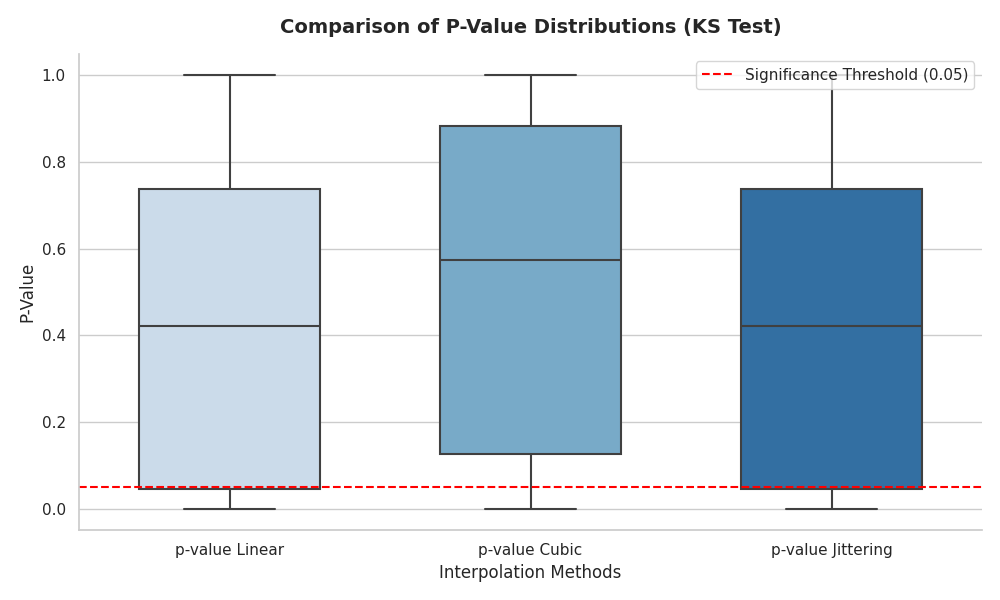}
\caption{Box plot of Kolmogorov--Smirnov test \textit{p-values}.}
\label{fig:KS_comparacao}
\end{figure}

Table \ref{tab:advanced_statistical_metrics} indicates superior performance for the \textit{Cubic Spline} method, which achieved lower distributional divergences (KL and JSD), better temporal alignment (lower mean DTW), and higher stability (smaller standard deviations). The infinite KL values observed for the \textit{Linear} and \textit{Jittering} methods arise from support mismatches in the KDE estimation (i.e., assigning zero probability to regions where the reference series has positive mass). The Kolmogorov--Smirnov test further supports the advantage of \textit{Cubic Spline}, reducing the proportion of municipalities with statistically significant differences relative to the reference weekly series (p$<0.05$), suggesting that in approximately 80\% of cases the interpolated series are statistically indistinguishable from the ground truth. In contrast, \textit{Jittering} did not yield consistent gains and may move the resulting distribution away from the observed epidemiological profile; Figure \ref{fig:KS_comparacao} reinforces this pattern, with \textit{Cubic Spline} exhibiting a higher median \textit{p-value}.

\begin{table}[htbp]
\centering
\caption{Distributional Similarity and Temporal Distance Metrics for Selected Municipalities}
\label{tab:advanced_municipal_metrics}
\resizebox{\textwidth}{!}{%
\begin{tabular}{@{}lcccccccccccc@{}}
\toprule
\textbf{Municipality} & \multicolumn{3}{c}{\textbf{KL Divergence} ($\downarrow$)} & \multicolumn{3}{c}{\textbf{JSD Distance} ($\downarrow$)} & \multicolumn{3}{c}{\textbf{DTW Distance} ($\downarrow$)} & \multicolumn{3}{c}{\textbf{p-value} ($\uparrow$)} \\ \cmidrule(lr){2-4} \cmidrule(lr){5-7} \cmidrule(lr){8-10} \cmidrule(lr){11-13}
 & \textit{Lin.} & \textit{Cub.} & \textit{Jit.} & \textit{Lin.} & \textit{Cub.} & \textit{Jit.} & \textit{Lin.} & \textit{Cub.} & \textit{Jit.} & \textit{Lin.} & \textit{Cub.} & \textit{Jit.} \\ \midrule
Arujá     & 0.002 & 0.009 & \textbf{0.001} & 0.024 & 0.047 & \textbf{0.017} & 368.90   & \textbf{211.05} & 376.21   & 0.293 & 0.293 & 0.293 \\
Braúna    & 0.419 & \textbf{0.114} & 0.367 & 0.341 & \textbf{0.180} & 0.320 & \textbf{14.90} & 16.50 & 15.19 & 0.002 & \textbf{0.004} & 0.002 \\
São Paulo & 0.018 & 0.032 & \textbf{0.009} & 0.066 & 0.090 & \textbf{0.049} & 102481.4 & \textbf{74075.8} & 94147.4 & 0.998 & 0.998 & 0.998 \\
Suzano    & \textbf{0.003} & 0.004 & 0.006 & \textbf{0.028} & 0.031 & 0.039 & 1287.15  & \textbf{685.82} & 1326.39  & 0.972 & \textbf{1.000} & 0.998 \\ \bottomrule
\end{tabular}%
}
\subcaption*{\small Note: ($\downarrow$) indicates that lower values are better; ($\uparrow$) indicates that higher values are better. KL: Kullback--Leibler; JSD: Jensen--Shannon; DTW: Dynamic Time Warping.}
\end{table}

In Table \ref{tab:advanced_municipal_metrics}, \textit{Jittering} achieves, in some cases, the lowest KL and JSD values (notably for Arujá and São Paulo), suggesting closer agreement between the estimated probability density functions of the interpolated and reference series. However, \textit{Cubic Spline} stands out in temporal alignment (DTW) and in the KS test, yielding the largest \textit{p-values}; for municipalities such as Suzano ($p \approx 1.000$), the statistical agreement is nearly complete. Therefore, for epidemiological surveillance applications, preserving the curve shape through cubic smoothing tends to be more informative than the rigidity of strictly linear methods.

\section{Dataset Description}

These dataset (\datasetversion) provides municipal time series of dengue-related hospitalizations in Brazil, harmonized and organized at weekly resolution. Weekly (epiweek) series are derived from the original monthly counts via a temporal disaggregation protocol based on \textit{cubic spline} interpolation with a correction step that, for the dengue target variable, strictly preserves the original monthly totals.

In addition to dengue hospitalization series, the dataset includes municipal covariates obtained from IBGE \cite{ibge2025}, such as population density, CH$_4$, CO$_2$, and NO$_2$ emissions, poverty and urbanization indices, maximum temperature, mean monthly precipitation, minimum relative humidity, as well as municipal latitude and longitude. These variables follow the same weekly indexing convention to support multivariate modeling.

The statistical and temporal fidelity of the weekly disaggregation was validated using a high-resolution reference dataset from the São Paulo State Epidemiological Surveillance Center (CVE-SP) for 2024 \cite{Dengue_cases_in_SP}, which reports both monthly and epidemiological-week counts. Based on this direct comparison, the interpolation strategy with the highest adherence was selected and subsequently applied to generate weekly series for the remaining Brazilian municipalities over 1999--2021.

The basic unit of observation is the dengue hospitalization rate associated with a municipality and an epidemiological week, accompanied by aligned municipal covariates. The dataset covers Brazilian municipalities ($n=5{,}028$) over the period 1999--2021 at weekly resolution.

\subsection{File Format and Package Structure}
The dataset package is organized under a top-level directory named \texttt{DengueDataset}, which includes documentation files and three data subdirectories: \texttt{data}, \texttt{target}, and \texttt{features}. The \texttt{README.md} file documents data provenance and preprocessing (DataSUS \cite{datasus2025} and IBGE \cite{ibge2025}), as well as the column schema, naming conventions, and usage instructions. The \texttt{LICENSE} file specifies the dataset license.

The \texttt{data/} directory contains one folder per Federative Unit (state), named according to the pattern \texttt{Dengue by state \{IBGE code\} - \{State name\}}. Each state folder includes a \texttt{.csv} file containing the target series (dengue hospitalization rate) and the corresponding covariates, all aligned at weekly resolution and interpolated using the recommended method (\textit{cubic spline}) over 1999--2021. This organization supports state-level stratified analyses and reduces I/O overhead when users require only geographic subsets.

The \texttt{target/} directory includes only the interpolated \textbf{dengue hospitalization rate} series, enabling direct use in univariate forecasting tasks and in workflows where covariates are optional. Finally, the \texttt{features/} directory provides one \texttt{.csv} file per explanatory variable interpolated in the \textit{cubic} format, including, for example, \texttt{Densidade\_demografica\_cubic.csv}, \texttt{Indice\_de\_urbanizacao\_cubic.csv}, \texttt{Indice\_de\_pobreza\_cubic.csv}, \texttt{Temperatura\_maxima\_cubic.csv}, \texttt{Precipitacao\_mensal\_media\_cubic.csv}, \texttt{Menor\_umidade\_relativa\_cubic.csv}, \texttt{Gee\_Co2\_cubic.csv}, \texttt{Gee\_No2\_cubic.csv}, and \texttt{GEE\_CH4\_cubic.csv}. This modular organization facilitates reuse of covariates and integration with other epidemiological targets.

\begin{table}[H]
\centering
\caption{File structure of the \texttt{DengueDataset} package (\datasetversion)}
\label{tab:files}
\resizebox{\textwidth}{!}{%
\begin{tabular}{@{}lll@{}}
\toprule
\textbf{Path} & \textbf{Description} & \textbf{Format} \\ \midrule
\texttt{DengueDataset/README.md} & Documentation (provenance, schema, usage) & Text \\
\texttt{DengueDataset/LICENSE} & Dataset license (\licensedata) & Text \\
\texttt{DengueDataset/data/} & State-level folders (\texttt{Dengue by state \{IBGE\} - \{State\}}) & Directory \\
\texttt{DengueDataset/data/.../*.csv} & Full state-level tables (target + covariates) at weekly resolution & CSV \\
\texttt{DengueDataset/target/} & Interpolated target series (dengue hospitalization rate) & Directory \\
\texttt{DengueDataset/target/*.csv} & Target-only files (fast/univariate use) & CSV \\
\texttt{DengueDataset/features/} & One table per interpolated covariate (\textit{cubic}) & Directory \\
\texttt{DengueDataset/features/*\_cubic.csv} & Covariate files (e.g., CO$_2$, NO$_2$, CH$_4$, climate, indices) & CSV \\ \bottomrule
\end{tabular}%
}
\end{table}

\subsection{License, Provenance, and Ethical Considerations}
The data is aggregated by municipality and period, containing no personal identifiers. Provenance includes DataSUS (monthly) and CVE-SP (2024 for validation) \cite{Dengue_cases_in_SP}.

\textbf{Dataset license:} \licensedata.

\textbf{Paper license on arXiv:} \licensepaper.

\section{Limitations and Considerations for Use}

This dataset provides weekly \textit{derived} time series obtained through temporal disaggregation of originally monthly data. Therefore, the released weekly resolution should not be interpreted as primary observations (i.e., it does not correspond to weekly records collected at the source for 1999--2021), but rather as a reconstruction constrained to be consistent with the original monthly totals. Below we summarize the main limitations and recommendations for responsible use.

\subsection{Derived nature of the weekly scale}
Disaggregation into epidemiological weeks is a deterministic/parameterized procedure whose fidelity depends on both the chosen interpolation strategy and the characteristics of each municipal series. Although the \textit{cubic spline} approach achieved the best statistical and temporal agreement during validation against the CVE-SP reference (2024), the resulting weekly series remains an approximation and may not fully reproduce genuine intra-month reporting patterns, such as reporting delays, weekend effects, or operational changes in surveillance workflows.

\subsection{Limitations under extreme sparsity}
In municipalities with very low incidence (series dominated by zeros and rare events), smooth methods such as \textit{cubic spline} may introduce artificial oscillations and intermediate values that do not correspond to actual occurrences, potentially affecting local metrics and the \textit{timing} of isolated outbreaks. In such settings, we recommend:
(i) considering the linear \textit{baseline} as a more conservative alternative;
(ii) applying minimum-incidence thresholds when including municipalities in weekly-scale analyses; and/or
(iii) adopting models designed for excessive zeros (e.g., \textit{zero-inflated} formulations).

\subsection{Validation limited to a specific reference dataset}
The recommended method was selected based on comparisons against the CVE-SP dataset (2024), which provides observed monthly and weekly counts for the state of São Paulo. While this reference enables direct and rigorous validation, generalization to other states and historical periods may be affected by regional and temporal differences, including:
(i) heterogeneity in notification processes and health-system coverage;
(ii) changes in coding and reporting practices over time; and
(iii) region-specific seasonality and transmission dynamics.
Accordingly, users focusing on specific regions are encouraged to perform sensitivity analyses (e.g., comparing multiple interpolation strategies or checking consistency with local sources when available).

\subsection{Calendar alignment assumptions}
Mapping civil months to epidemiological weeks necessarily involves approximations, since epidemiological weeks may cross month boundaries. Although the protocol preserves the number of weeks associated with each month and, for dengue, enforces conservation of monthly totals, small alignment discrepancies may arise when directly comparing monthly and weekly series. For calendar-sensitive tasks (e.g., outbreak onset detection), we recommend explicitly stating the adopted convention and, whenever feasible, using native epidemiological calendars.

\subsection{Covariates: temporal propagation and interpretation}
The provided covariates (e.g., population density, urbanization, poverty indices, emissions, and climate variables) are not additive quantities and therefore do not follow the monthly-sum conservation rule applied to the dengue target. At weekly resolution, these covariates were aligned by propagating the monthly (or annual) values to the corresponding epidemiological weeks to ensure temporal consistency with the target series. Consequently, intra-month variability of covariates is not represented, and weekly-scale models may attribute explanatory power to fluctuations that, in practice, are concentrated in the interpolated target. We recommend:
(i) interpreting covariate effects primarily at monthly/seasonal scales;
(ii) considering lags and smoothing operations; and
(iii) avoiding causal inference without external validation.

\subsection{Implications for modeling and evaluation}
Because the weekly series are reconstructed, weekly evaluation metrics may overestimate performance when models learn patterns induced by interpolation, particularly in low-variability settings.

In summary, this dataset is appropriate for epidemiological forecasting and spatiotemporal modeling provided that the derived nature of the weekly scale is explicitly considered. We recommend using the \textit{cubic spline} method as the default, while applying sensitivity analyses and conservative alternatives for municipalities with extreme sparsity, where linear approaches may be more robust.

\section{Conclusion}

This data paper introduced a (\datasetversion) version, a public and versioned dataset (Zenodo, DOI \href{\dataseturl}{\datasetdoi}) that organizes municipal dengue hospitalization time series in Brazil at a weekly resolution obtained via temporal disaggregation. In addition to the epidemiological target, the package provides explanatory variables commonly used in epidemiological and environmental modeling (e.g., population density, socioeconomic indices, climate indicators, and emissions), harmonized to support multivariate analyses and predictive modeling.

The primary motivation of the dataset is to overcome the limited sample length at the monthly scale (264 observations per municipality from 1999 to 2021) by increasing temporal granularity to epidemiological weeks, thereby enabling more effective training of \textit{Machine Learning} models with higher predictive capacity \cite{melhor_perfomasse}. To ensure that this temporal expansion does not introduce undue distortions, we rigorously evaluated the statistical and temporal adherence of three interpolation strategies (\textit{Linear}, \textit{Jittering}, and \textit{Cubic Spline}) using as reference the 2024 dataset from the São Paulo State Epidemiological Surveillance Center (CVE-SP) \cite{Casos_dengue_SP}, which provides observed counts at both monthly and epidemiological-week resolutions, allowing direct comparison.

The results showed that, although traditional error metrics (MAE and RMSE) were broadly competitive across methods, \textit{Cubic Spline} interpolation performed better in higher-incidence scenarios by more effectively capturing outbreak acceleration and deceleration phases. The selection of the recommended approach was primarily supported by distributional similarity and temporal-distance criteria (KL, JSD, DTW, and the KS test), under which cubic smoothing achieved the highest morphological and statistical fidelity to the reference series. Accordingly, \textit{Cubic Spline} was adopted as the default method for generating the weekly series released in the dataset.

An important caveat should be noted: in municipalities with extreme sparsity (series dominated by zeros and rare events), \textit{Cubic Spline} may introduce artificial polynomial oscillations, as observed in representative cases (e.g., Braúna). In such contexts, the linear \textit{baseline} tends to be more conservative and robust, and we recommend that users perform sensitivity analyses, particularly when the objective involves local inference or precise detection of rare events.

Overall, these dataset provides a standardized and reproducible resource for epidemiological forecasting and spatiotemporal modeling by offering (i) original monthly series, (ii) weekly derived series validated against an observed reference, and (iii) temporally aligned covariates for multivariate use. As future work, we plan to incorporate advanced time-series \textit{Data Augmentation} techniques (e.g., \textit{Scaling}, \textit{Rotation}, and \textit{Time Warping}) to increase the robustness of deep learning models and improve generalization across population scales and regional contexts. In addition, we aim to expand this initiative by creating and curating further public-health-oriented epidemiological datasets covering other diseases and syndromes of interest, thereby supporting surveillance, decision-making, and evidence-based intervention planning.

\bibliographystyle{ieeetr}
\bibliography{references}

\end{document}